\documentclass[11pt]{article}
\usepackage{geometry}
\usepackage{coling2020}
\usepackage{times}
\usepackage{url}
\usepackage{latexsym}
\usepackage{microtype}

\usepackage{graphicx}
\usepackage{wrapfig}
\usepackage{caption}
\usepackage{subcaption}

\usepackage{enumitem}
\usepackage{amsmath}
\usepackage{array}
\usepackage{color}

\usepackage{soul}

\colingfinalcopy 

\title{CS-NET at SemEval-2020 Task 4: Siamese BERT for ComVE}

 \author{Soumya Ranjan Dash$^{*}$ \qquad    
  Sandeep Routray$^{*}$  \qquad  
  Prateek Varshney\thanks{\quad Authors equally contributed  to this work.} \qquad
  Ashutosh Modi  \\
{Indian Institute of Technology Kanpur (IITK)} \\
  {\tt \{soumyard, sroutray\}@iitk.ac.in, varshney@cse.iitk.ac.in}  \\
  {\tt ashutoshm@cse.iitk.ac.in}  \\
}

\date{}

\begin{document}
\maketitle

\begin{abstract}
  In this paper, we describe our system for Task 4 of SemEval 2020, which involves differentiating between natural language statements that confirm to common sense and those that do not. The organizers propose three subtasks - first, selecting between two sentences, the one which is against common sense. Second, identifying the most crucial reason why a statement does not make sense. Third, generating novel reasons for explaining the against common sense statement. Out of the three subtasks, this paper reports the system description of subtask A and subtask B. This paper proposes a model based on transformer neural network architecture for addressing the subtasks. The novelty in work lies in the architecture design, which handles the logical implication of contradicting statements and simultaneous information extraction from both sentences. We use a parallel instance of transformers, which is responsible for a boost in the performance. We achieved an accuracy of 94.8\% in subtask A and 89\% in subtask B on the test set.
\end{abstract}

\section{Introduction}
\label{intro}

%
%
\blfootnote{
    %
    %
    %
    %
    %
    %
    \hspace{-0.65cm}  
    This work is licensed under a Creative Commons 
    Attribution 4.0 International License.
    License details:
    \url{http://creativecommons.org/licenses/by/4.0/}.
}

Incorporating common sense in natural language understanding systems and evaluating whether a system has sense-making capability remains a fundamental question in the natural language processing field \cite{Modi17,modi:CONLL2016,modi:CONLL2014}. One important difference between human and machine text understanding lies in the fact that humans have access to commonsense knowledge while processing text, which helps them to draw inferences about facts that are not mentioned in a text, but that is assumed to be common ground \cite{TACL968}. For a  computer system,  inferring unmentioned facts is a non-trivial challenge \cite{ostermann2018mcscript}. For our problem, we have proposed methods to include common sense in the validation and reasoning paradigm \cite{yang}.

\begin{figure}
     \centering
     \includegraphics[width=\textwidth]{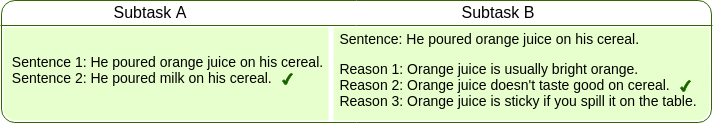}
     \caption{Sample data point explaining the problem}
     \vspace{-3mm}
     \label{fig:eg}
\end{figure}

Task 4 of semeval 2020 \cite{wangsemeval} is a common-sense validation and explanation task. It consists of classifying against common sense sentences from sentences that make sense. Figure \ref{fig:eg} shows examples from subtask A and subtask B. In subtask A, clearly sentence 1 is against common sense. Subtask B contains three options for reasons to explain why sentence 1 is against common sense. As orange juice does not taste good on cereal, but milk does, sentence 1 makes less sense than sentence 2.

We use the generated embedding from transformer based encoders like BERT \cite{bert}, RoBERTa \cite{liu2019roberta}, AlBERT \cite{lan2019albert} which capture in-context semantic information and design a siamese architecture to extract the relational information between the sentences. The implementation for our system is made available via Github\footnote{\url{https://github.com/soumyardash/SemEval2020-Task4}}.


\section{Background}
\subsection{Problem Definition}
The task is to develop a system that can differentiate natural language statements that make sense from those that do not. The organizers present three subtasks to test these. Out of these, we describe the description of the system proposed for the first two subtasks, subtask A and subtask B. Each instance in the given dataset is composed of 10 sentences: $\{S_1, S_2, O_1, O_2, O_3, R_1, R_2, R_3\}$. $S_1$ and $S_2$ are two syntactically similar statements, differing by only a few words, with only one making sense.
\begin{itemize}
    \item \textbf{Subtask A \textit{Validation}:} Requires the model to choose which of the two statements $S_1$ and $S_2$ does not makes sense. We frame it as a binary classification problem and estimate the probability that the sentence is against common sense. Accuracy score is used for evaluation.
    \item \textbf{Subtask B \textit{Explanation (Multi-Choice)}:} Requires the model to choose the most appropriate of the three reasons $\{O_1, O_2, O_3\}$ to explain the against-common-sense statement (one of $S_1$ or $S_2$). We formulate this as a multi-class classification problem and estimate the probability that the reason is in fact the correct explanation. Accuracy score is used for evaluation.
\end{itemize}

\subsection{Related Work}
\textbf{Language models :}
Over the years, a great amount of effort has been directed towards creating benchmark datasets that can measure a system's performance on language processing tasks and provide an impetus for the development of new approaches to the tasks. These benchmark tasks have led to many computational models ranging from earlier symbolic and statistical approaches to recent approaches based on deep neural networks; which model context of language, take advantage of external data or knowledge resources, and achieve the state-of-the-art performance, and at times, even near or above human performance. A major landmark in NLP is the development of pre-trained models and embeddings that can be used as features or further fine-tuned for downstream tasks. These models are often trained based on large corpora of textual data to capture different word senses.

The defining contribution of Embeddings from Language Models (ELMo) \cite{peters2018deep} is its contextual word embeddings, which are built relying on the entire input sentence that they belong to. The recent  Bidirectional  Encoder  Representations from  Transformers  (BERT)  model outperforms previous competitive approaches by better capturing the context. A  more recent model,  XLNET \cite{yang2019xlnet},  exceeded the performance of the vanilla BERT variant on several benchmarks. Robustly Optimized BERT Approach (RoBERTa) \cite{liu2019roberta} achieved further improvement by making changes to the pre-training approach used in BERT. It includes randomizing masked tokens in the cloze pre-training task for each epoch instead of keeping them the same over epochs. It also augments the next sentence prediction pre-training task with an additional task which compels the model to also predict whether a candidate next sentence comes from the same document or not. A Lite BERT (ALBERT) \cite{lan2019albert} implements several novel parameter reduction techniques to increase the training speed and efficiency of BERT, enabling a much deeper scale-up than the original large variant of BERT while having fewer parameters. ELECTRA \cite{clark2020electra} is used to pre-train transformers with comparatively less computation. This model is similar to the discriminator of a GAN.

\textbf{Common sense validation :}
TriAN \cite{wang2018yuanfudao} achieved state-of-the-art performance   for   SemEval   ’18   Task-11 : Machine Comprehension Using Commonsense Knowledge \cite{ostermann-etal-2018-semeval}. It proposed a threeway attention mechanism to model interactions between the text, question, and answers, on top of BiLSTMs. It incorporated relational features (based on ConceptNet).

CommonsenseQA \cite{talmor2018commonsenseqa} is a large multi-choice question answering dataset that requires different types of commonsense knowledge to predict the correct answers. The leaderboard provides a variety of ensembling techniques using different LMs and also several instances of the same LM.


\section{System Overview}
We mention some crucial insights from our initial experimentation that were helpful to arrive at the final approach. We consider both subtasks separately.
\subsection{Subtask A: Validation}
\subsubsection{Initial Experimentation}
Initially, we experimented with the vanilla implementation of BERT sequence classifier model. It consists of a linear classification head over the BERT embedding of the [CLS] token which are supposed to have learnt representation for the entire sentence. The classifier predicts a binary label indicating if the input sentence made sense or not. The weights of the BERT encoder stack are fixed in this approach. Even though the accuracy obtained (see Table\ref{tab:stA_perf}) was higher than the baseline, it was still on the lower side. 

Improving upon the initial approach, we then updated the weights of BERT encoder during training phase. We also modified the training samples as - if $S_1$ and $S_2$ are the given sentence pair, we concatenated these with the help of a phrase, e.g. $S_1$ \textit{makes more sense than} $S_2$, and $S_2$ \textit{makes more sense than} $S_1$. The task was now to predict binary label depending on the correctness of the claim. These changes resulted in a significant gain in the performance. Interestingly, we observed that the performance varied slightly depending on the conjunctive phrase used to join the sentences. While conjunctive phrases like "makes more sense than" gave better results, the phrases like "because" and "not valid" had a lower performance. Further, we concluded that training the BERT encoder was necessary to achieve good performance.

We also observed that the approach proposed above suffered from logical fallacy, that is, in case of mispredictions the model assigned both sentences of the pair as against commonsense. Further, though the approach worked well in the trainset, it did not generalize well to the devset, indicating overfitting. 

\subsubsection{Proposed Approach}
\begin{figure}
     \centering
     \begin{subfigure}[b]{0.45\textwidth}
         \centering
         \includegraphics[width=0.8\textwidth]{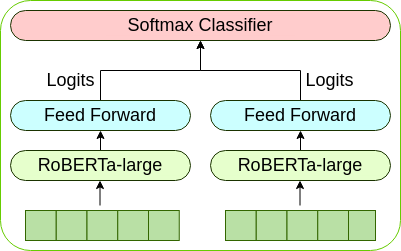}
         \caption{Validation Subtask}
         \label{fig:stA}
     \end{subfigure}
     \hfill
     \begin{subfigure}[b]{0.45\textwidth}
         \centering
         \includegraphics[width=\textwidth]{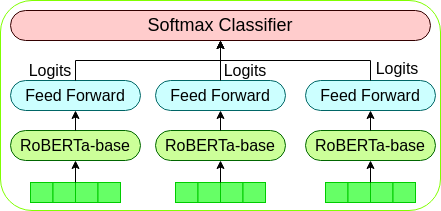}
         \caption{Multi-choice explanation Subtask}
         \label{fig:stB}
     \end{subfigure}
     \caption{Final architecture used for the subtasks}
     \label{fig:final}
\end{figure}

To address the issues discussed above, we designed a siamese architecture (see Fig.\ref{fig:stA}). The model took both the sentence at the same time for prediction and predicted a probability score over the two sentences.
Mathematically,
\begin{equation*}
    x_1 = E(s_1) \qquad x_2 = E(s_2) \qquad l_1 = \frac{e^{Wx_1}}{e^{Wx_1}+e^{Wx_2}} \qquad l_2 = \frac{e^{Wx_2}}{e^{Wx_1}+e^{Wx_2}}
\end{equation*}
where $s_1$ and $s_2$ are vectorized representations of tokens of $S_1$ and $S_2$. $E(.)$ denotes the trainable transformer encoder block weights and $W$ is the trainable classifier weights. $x_1$ and $x_2$ are the representations of the starting [CLS] token. Cross-entropy loss was used while training to determine the optimal weights.

Note that $E(.)$ and $W$ are same for both the sentences $S1$ and $S2$. Since, the weights off the two networks (transformer based encoder+classifier) were shared, the model was able to focus better on the features which differed between the two sentences and avoid overfitting. Such features proved crucial in deciding the output label which was supported by the superior performance on both trainset and devset. We also experimented with variants of BERT: BERT-base, BERT-large, RoBERTa-base and RoBERTa-large. The best performance was obtained with RoBERTa-large, which formed our final submission. We did not observe any significant gains in using conjunctive phrases and hence excluded it from our final submission.

\subsection{Subtask B: Multi-Choice}
\subsubsection{Initial Experimentation}
As an initial approach, we formulated the task as a binary classification problem using the BERT sequence classifier model. Using the given against commonsense sentence $S$ from the dataset and the three candidate reasons $O_i, i=1,2,3$, we formed sentences $S$ \textit{makes less sense because }$O_i$. Although the performance obtained using this approach was above the baseline, it was still on the lower side (see Table\ref{tab:stB_perf}).

Next, we tried using conjunctive phrases. We formed \textit{The reason }$S$ \textit{makes less sense is} $O_1$ \textit{rather than} $O_2$ \textit{or} $O_3$, with all possible permutations of $O_1,O_2,O_3$ and used it as our input instead of original sentences. The idea was to allow the model to access information from other candidate reasons while deciding the correctness of a particular reason. This approach gave a good performance on the trainset but did not perform well on the devset.

Above approaches also suffered from similar logical fallacy, as described in Section 3.1.1. Motivated by the success of siamese architecture in avoiding over-fitting, enabling efficient information sharing, and eliminating logical fallacies, we designed a three-way siamese architecture for subtask B (See Fig.\ref{fig:stB}).

\subsubsection{Proposed Approach}
First, we formed the three inputs by joining $S$ with $O_1,O_2, O_3$ respectively using a separator token ([SEP]). We did not use any conjunctive phrases in our final approach. The model took the inputs at the same time and predicted a probability score indicating the likelihood of the reason being the correct explanation. Mathematically,
\begin{align*}
    x_1 &= E(s_1) & x_2 &= E(s_2) & x_3 &= E(s_3) \\
    l_1 &= \frac{e^{Wx_1}}{\Sigma} & l_2 &= \frac{e^{Wx_2}}{\Sigma} & l_3 &= \frac{e^{Wx_3}}{\Sigma}
\end{align*}
where $s_1, s_2, s_3$ are the vectorized representation of input tokens and $\Sigma=e^{Wx_1}+e^{Wx_2}+e^{Wx_3}$.  $E(.)$ denotes the trainable transformer encoder block weights and $W$ is the the trainable classifier weights. Cross-entropy loss function was used while training to determine the optimal weights.

The shared weight among the inputs enables efficient information exchange. The model actually compares the reasons against each other while determining the correct explanation. We experimented with using the [CLS] token representation and an average pooled representation of all the tokens as the input to feed-forward classification layer. We observed that better performance was obtained with an average pooling. The performance gains can be attributed to the fact that average pooling preserved more useful information for the classification layer, which was otherwise lost. We also experimented with variants of BERT: BERT-base, BERT-base, AlBERT-base, and RoBERTa-base.  The best performance was obtained with RoBERTa-base and average pooling, which formed our final submission.

\section{Experimental Setup}
\textbf{Data:} Dataset that was used to build the models were provided by the organizers as a part of the pilot study \cite{data}. As a single datapoint, Subtask A had an against-common-sense and a correct sentence. For each of the against common sense sentence of Subtask A, Subtask B had a sensible reason and two confusing reasons. The datasets were provided in three phases. The training data was used to train the language model. It consisted of 10,000 datapoints. The 997 datapoints long development dataset was used to tune the hyperparameters of the language model. The answers generated on the test dataset of 1,000 datapoints were used for submission. From Table \ref{tab:dataset}, we observe that the label distribution is unbiased for training as well as the development dataset.  


\begin{table}
\small
    \begin{subtable}[h]{0.4\textwidth}
        \centering
        \begin{tabular}{|m{0.3\textwidth}|m{0.27\textwidth}|m{0.33\textwidth}|}     
            \hline 
              Dataset & First sentence invalid & Second sentence invalid \\
             \hline
             Training Data & 4979 & 5021 \\ 
             Validation Data & 518 & 479 \\ 
             \hline
        \end{tabular}
       \caption{Subtask A dataset details}
       \label{tab:stA_data}
    \end{subtable}
    \hfill
    \begin{subtable}[h]{0.55\textwidth}
        \centering
        \begin{tabular}{|m{0.22\textwidth}|m{0.18\textwidth}|m{0.22\textwidth}|m{0.18\textwidth}|} \hline
              Dataset & First reason correct & Second reason correct & Third reason correct \\ \hline
             Training Data & 3195 & 3362 & 3443 \\ 
             Validation Data & 344 & 327 & 336\\ \hline
        \end{tabular}
        \caption{Subtask B dataset details}
        \label{tab:stB_data}
     \end{subtable}
     \caption{Dataset analysis}
     \label{tab:dataset}
\vspace{-2mm}
\end{table}

\textbf{Parameter setting:} We used held-out validation using the development dataset for validation. The train:dev ratio for the training was nearly 10:1. The hyperparameters were tuned using a grid search around the default setting of the language model. It ensures the highest efficiency of the model. For both the subtasks, the accuracy of the model on the development dataset was used as a measure of performance. Our submitted model is trained on batch mode with a batch size of 32 using AdamW optimizer. The learning rate of the optimizer was set to be 2e-5, and the adam epsilon was set to be 1e-8.

\section{Results}
\subsection{Result Analysis}
\begin{table}
\small
    \begin{subtable}[h]{0.4\textwidth}
        \centering
        \begin{tabular}{|c|c|}     
            \hline 
             Model & Accuracy \\ 
             \hline
             BERT Classifier & 77.1\% \\ 
             BERT Classifier + phrase concat. & 84.3\% \\ 
             Albert-base Siamese & 87.6\% \\
             BERT-base Siamese & 88.6\% \\ 
             RoBERTa-base Siamese & 90.7\% \\
             Electra-base Siamese & 93.6\% \\
             RoBERTa-large Siamese & 95.2\% \\ 
             \hline
        \end{tabular}
       \caption{Subtask A accuracy on dev set}
       \label{tab:stA_perf}
    \end{subtable}
    \hfill
    \begin{subtable}[h]{0.5\textwidth}
        \centering
        \begin{tabular}{|c|c|} \hline
             Model & Accuracy \\ \hline
             BERT Classifier & 77.3\% \\ 
             BERT Classifier + phrase concat & 83.2\% \\ 
             BERT-base Siamese & 84.3\% \\ 
             AlBERT-base Siamese & 85.7\% \\
             RoBERTa-base Siamese & 87.5\% \\ 
             Electra-base Siamese & 87.7\% \\
             RoBERTa-base Siamese+avg. pool & 89.7\% \\ \hline
        \end{tabular}
        \caption{Subtask B accuracy on dev set}
        \label{tab:stB_perf}
     \end{subtable}
     \caption{Results analysis of various models}
     \label{tab:performance}
\vspace{-2mm}
\end{table}

Table \ref{tab:performance} compiles the system performance on the development dataset for various approaches.

For subtask A, Table \ref{tab:stA_perf} demonstrates the effect of adding conjunctive phrases to the sentence pair over the vanilla implementation of BERT. It exhibits the gain in accuracy obtained from addressing the logical fallacy problem by using a two-way siamese network, as discussed previously. It compares the performance of various models, notable among which is Electra and RoBERTa, which outperform BERT. 

Table \ref{tab:stB_perf} exhibits a similar trend for subtask B, with a gain in accuracy after the use of concatenating phrases with the sentence and all its reasons. The use of the three-way siamese network further improves performance. The table also shows the effect of different language models on the task when used with the siamese architecture. The addition of an average pooling layer over the language model embedding gave improved results in comparison to the use of embedding of the [CLS] token only. 

\subsection{Error Analysis}
For subtasks A and B, we found our system to be performing well on leader-board. Our system ranked 10th in subtask A and 12th in subtask B. While our system gave an accuracy of 94.8\% on test set in subtask A, the top performance was 97\%. In subtask B, the accuracy of our system was 89\%, the top performance was 95\%. There was a great diversity of contexts in our dataset. Our model learnt representations to capture the context. We believe the performance could be further improved by directly considering the relationship established between the keywords in the sentences. One way to do this would be to encode word relationships in a sentence using a knowledge base like ConceptNet. Then an attention layer could be used to learn a joint representation using representations from BERT-variant and ConceptNet.

\section{Conclusion}
In this paper, we have presented our systems for the Commonsense Validation and Explanation Challenge in SemEval2020. Our approach for subtask A and subtask B achieves close to state-of-the-art results. Our approach demonstrates the advantage of using parallel instances of the transformer in terms of a performance gain in classification based tasks. 
\bibliographystyle{coling}
\bibliography{semeval2020}

\end{document}